\documentclass[twoside]{article}
\usepackage[accepted]{aistats2016}

\RequirePackage{fix-cm}

\usepackage{times}
\usepackage{wrapfig}
\usepackage{amstext}

\usepackage{dsfont}
\usepackage{hyperref}
\usepackage{url}
\usepackage{graphicx,times,amsmath} 
\usepackage{url}
\usepackage{graphicx}
\usepackage{amsfonts}
\usepackage{wasysym}
\usepackage{latexsym}
\usepackage{multirow}
\usepackage{amssymb}
\usepackage{epstopdf}
\usepackage{times}
\usepackage{wrapfig}
\usepackage{natbib}

\usepackage{array}
\usepackage{multirow} 
\usepackage{verbatim}
\usepackage{amstext}
\usepackage{rotating}
\usepackage{graphicx}
\usepackage{subfigure}
\usepackage{multirow}
\usepackage{url}
\usepackage{colortbl}
\usepackage{stmaryrd}
\usepackage{appendix}

\usepackage{algorithm}
\usepackage{algorithmic}

\usepackage{amssymb,amsmath,amsthm, amsfonts}

\begin{document}

\twocolumn [
\aistatstitle  {Random Forest for the Contextual Bandit Problem }

\aistatsauthor{ Rapha\"{e}l F\'{e}raud \And Robin Allesiardo \And Tanguy Urvoy \And Fabrice Cl\'{e}rot}

\aistatsaddress{raphael.feraud@orange.com \\Orange Labs, France\And robin.allesiardo@orange.com \\Orange Labs, France\And tanguy.urvoy@orange.com\\Orange Labs, France \And fabrice.clerot@orange.com \\
Orange Labs, France} ]

\begin{abstract}

To address the contextual bandit problem, we propose an online {\it random forest} algorithm.
The analysis of the proposed algorithm is based on the sample complexity needed to find the optimal decision stump. 
Then, the decision stumps are recursively stacked in a random collection of decision trees, {\sc Bandit Forest}. 
We show that the proposed algorithm is near optimal.
The dependence of the sample complexity upon the number of contextual variables is logarithmic.
The computational cost of the proposed algorithm with respect to the time horizon is linear.
These analytical results allow the proposed algorithm to be efficient in real applications, where the number of events to process is huge, 
and where we expect that some contextual variables, chosen from a large set, have potentially non-linear dependencies with the rewards. 
In the experiments done to illustrate the theoretical analysis, {\sc Bandit Forest} obtain promising results in comparison with 
state-of-the-art algorithms. 
\end{abstract}

\section{Introduction}

By interacting with streams of events, machine learning algorithms are used for instance to optimize the choice of ads on a website, to choose the best human machine interface, 
to recommend products on a web shop, to insure self-care of set top boxes, to assign the best wireless network to mobile phones.
With the now rising {\it internet of things}, the number of decisions (or actions) to be taken by more and more autonomous devices further increases.  
In order to control the cost and the potential risk to deploy a lot of machine learning algorithms in the long run, we need scalable algorithms which provide strong 
theoretical guarantees.

Most of these applications necessitate to take and optimize decisions with a partial feedback. 
Only the reward of the chosen decision is known. Does the user click on the proposed ad ? The most relevant ad is never revealed. Only the click on the proposed ad is known.
This well known problem is called {\it multi-armed bandits} ({\sc mab}). In its most basic formulation, it can be stated as follows:
there are $K$ decisions, each having an unknown distribution of bounded rewards. At each step, one has to choose a decision and receives a reward.
The performance of a {\sc mab} algorithm is assessed in terms of {\it regret} (or opportunity loss) with regards to the unknown optimal decision.
Optimal solutions have been proposed to solve this problem  using a stochastic formulation in \cite{ABF02}, using a Bayesian formulation in \cite {KKM12}, or using an adversarial formulation in \cite {ABFS02}. While these approaches focus on the minimization of the expected regret, the PAC setting (see \cite {V84}) or the $(\epsilon,\delta)$-best-arm identification, focuses on the sample complexity (i.e. the number of time steps) needed to find an $\epsilon$-approximation of the best arm with a failure probability of $\delta$. This formulation has been studied for {\sc mab} problem in \cite {EMM02,BMS09}, for dueling bandit problem in \cite {UCFN13},  and for linear bandit problem in \cite {SLM14}. 

Several variations of the {\sc mab} problem have been introduced in order to fit practical constraints coming from ad serving or marketing optimization. These variations include for instance the death and birth of arms in \cite {CKRU08}, the availability of actions in \cite {KNS08} or a drawing without replacement in \cite {FU12,FU13}. But more importantly, in most of these applications, a rich contextual information is also available. 
For instance, in ad-serving optimization we know the web page, the position in the web page, or the profile of the user.
This contextual information must be exploited in order to decide which ad is the most relevant to display. 
Following  \cite {LZ07,DHKKLRZ11}, in order to analyze the proposed algorithms, we formalize below the {\it contextual bandit problem} (see Algorithm \ref {OCDP}).

Let ${\bf x_t} \in \{0,1\}^M$ be a vector of binary values describing the environment at time $t$.
Let ${\bf y_t} \in [0,1]^K$ be a vector of bounded rewards at time $t$, and $y_{k_t}(t)$ be the reward of the action (decision) $k_t$ at time $t$. 
Let $D_{x,y}$ be a joint distribution on $(\bf x,\bf y)$.  
Let $\mathcal{A}$ be a set of $K$ actions. Let $\pi:  \{0,1\}^M \rightarrow \mathcal{A}$ be a policy, and $\Pi$ the set of policies.
 
\begin{algorithm}[htb]
   \caption{The contextual bandit problem}
   \label{OCDP}
\begin{algorithmic}
	\REPEAT
	\STATE $({\bf x_t},{\bf y_t})$ is drawn according to $D_{x,y}$
	\STATE ${\bf x_t}$ is revealed to the player
	\STATE The player chooses an action $k_t = \pi_t({\bf x_t})$
	\STATE The reward $y_{k_t}(t)$ is revealed
	\STATE The player updates its policy $\pi_t$
	\STATE $t=t+1$
\UNTIL {$t=T$}
\end{algorithmic}
\end{algorithm}

Notice that this setting can easily be extended to categorical variables through a binary encoding.
The optimal policy $\pi^*$ maximizes the expected gain:

\begin {displaymath}
\pi^* = \arg \max_{\pi \in \Pi}  \mathds{E}_{D_{x,y}} \left[ y_{\pi({\bf x_t})}\right] 
\end {displaymath}

Let $k^*_t=\pi^*({\bf x_t})$ be the action chosen by the optimal policy at time $t$. 
The performance of the player policy is assessed in terms of expectation of the accumulated regret against the optimal policy with respect to $D_{x,y}$:
\begin {displaymath}
 \mathds{E}_{D_{x,y}} \left[R(T) \right] = \sum_{t=1}^T  \mathds{E}_{D_{x,y}} \left[y_{k^*_t}(t) - y_{k_t}(t) \right] \text { ,}
\end {displaymath}
where $R(T)$ is the accumulated regret at time horizon $T$.

\section{Our contribution} 
Decision trees work by partitioning the input space in hyper-boxes. They can be seen as a combination of rules, where only one rule is selected for a given input vector. Finding the optimal tree structure (i.e. the optimal combination of rules) is NP-hard.
For this reason, a greedy approach is used to build the decision trees offline (see \cite {B84}) or online (see \cite {DH00}).
The key concept behind the greedy decision trees is the decision stump.
While Monte-Carlo Tree Search approaches (see \cite {KS06}) focus on the regret minimization (i.e. maximization of gains), 
the analysis of proposed algorithms is based on the sample complexity needed to find the optimal decision stump with high probability.
This formalization facilitates the analysis of decision tree algorithms. Indeed, to build a decision tree under limited resources, 
one needs to eliminate most of possible branches. The sample complexity is the stopping criterion we need to stop exploration of unpromising branches.
In {\sc Bandit Forest}, the decision stumps are recursively stacked in a random collection of $L$ decision trees of maximum depth $D$. 
We show that {\sc Bandit Forest} algorithm is near optimal with respect to a strong reference: a {\it random forest} built knowing the joint distribution of the contexts and rewards.

In comparison to algorithms based on search of the best policy from a finite set of policies (see \cite {ABFS02,DHKKLRZ11,AHKLLS14}) our approach has several advantages.
First, we take advantage of the fact that we know the structure of the set of policies to obtain a linear computational cost 
with respect to the time horizon $T$. 
Second, as our approach does not need to store a weight for each possible tree, we can use deeper rules without exceeding the memory resources.
In comparison to the approaches based on a linear model (see {\sc LinUCB} in \cite {LCLS10}), our approach also has several advantages. First, it is better suited for the case where the dependence between the rewards and the contexts is not linear. Second, the dependence of regret bounds of the proposed algorithms on the number of contextual variables is in the order of $O(\log M)$ while the one of linear bandits is in $O(\sqrt {M})$ \footnote {In fact the dependence on the number of contextual variables of the gap dependent regret bound is in $O(M^2)$ (see Theorem 5 in \cite {APS11}).}.
Third, its computational cost with respect to time horizon in $O(LMDT)$ allows to process large set of variables, while linear bandits are penalized by the update of a $M \times M$ matrix at each update, which leads to a computational cost in $O(KM^2T)$.

\section {The decision stump}

In this section, we consider a model which consists of a decision stump based on the values
of a single contextual variable, chosen from at set of $M$ binary variables.

\subsection {A gentle start}

In order to explain the principle and to introduce the notations, before describing the decision stump used to build {\sc Bandit Forest}, we illustrate our approach on a toy problem. 
Let $k_1$ and $k_2$ be two actions. Let $x_{i_1}$ and $x_{i_2}$ be two binary random variables, describing the context. 
In this illustrative example, the contextual variables are assumed to be independent.
Relevant probabilities and rewards are summarized in Table \ref {tab1}. $\mu_{k_2}^{i_1} | v$ denotes the conditional expected reward of the action $k_2$ given $x_{i_1}=v$, and $P(x_{i_1}=v)$ denotes the probability to observe $x_{i_1}=v$ .

\begin{table} [htb]
\caption{The mean reward of actions $k_1$ and $k_2$ knowing each context value, and the probability to observe this context.}
\label{tab1}       
\begin {center}
\begin{tabular}{c|cc}
 & $v_0$ & $v_1$ \\
\hline
$\mu_{k_1}^{i_1}|v$ & $0$ & $1$\\
$\mu_{k_2}^{i_1}|v$ & $3/5$ & $1/6$\\[+0.1em]
\hline
$P(x_{i_1}=v)$ & $5/8$ & $3/8$ \\
\hline
\hline
$\mu_{k_1}^{i_2}|v$ & $1/4$ & $3/4$ \\ 
$\mu_{k_2}^{i_2}|v$ & $9/24$ & $5/8$ \\[+0.1em]
\hline
$P(x_{i_2}=v)$ & $3/4$ & $1/4$ \\
\hline
\end{tabular}
\end {center}
\end{table}

We compare the strategies of different players.
Player~$1$ only uses uncontextual expected rewards, while Player~$2$ uses the knowledge of $x_{i_1}$ to decide.
According to Table~\ref {tab1}, the best strategy for Player~$1$ is to always choose action $k_2$. His expected reward will be $\mu_{k_2}=7/16$.
Player $2$ is able to adapt his strategy to the context: his best strategy is to choose $k_2$ when $x_{i_1}=v_0$ and $k_1$ when $x_{i_1}=v_1$. According to Table~\ref{tab1}, his expected reward will be:
\begin {displaymath} 
\begin {split}
\mu^{i_1}  & =  P(x_{i_1}=v_0) \cdot \mu_{k_2}^{i_1} | v_0 +  P(x_{i_1}=v_1) \cdot \mu_{k_1}^{i_1} | v_1  \\
& = \mu_{k_2,v_0}^{i_1} + \mu_{k_1,v_1}^{i_1} =3/4 \text { ,}
\end {split}
\end {displaymath} 
where $\mu_{k_2,v_0}^{i_1}$ and $\mu_{k_1,v_1}^{i_1}$ denote respectively the expected reward of the action $k_2$ and $x_{i_1}=v_0$, and the expected reward of the action $k_1$ and $x_{i_1}=v_1$.
Whatever the expected rewards of each value, the player, who uses this knowledge, is the expected winner. 
Indeed, we have:
\begin {displaymath} 
\mu^i = \max_k \mu_{k,v_0}^i + \max_k \mu_{k,v_1}^i \geq  \max_k \mu_{k}
\end {displaymath} 
Now, if a third player uses the knowledge of the contextual variable $x_{i_2}$, his expected reward will be:
\begin {displaymath} 
\mu^{i_2}  = \mu_{k_2,v_0}^{i_2} + \mu_{k_1,v_1}^{i_2}=15/32
\end {displaymath} 
Player $2$ remains the expected winner, and therefore $x_{i_1}$ is the best contextual variable to decide between $k_1$ and $k_2$. 

The best contextual variable is the one which maximizes the expected reward of best actions for each of its values. We use this principle to build a reward-maximizing decision stump.

\subsection {Variable selection}

Let $\mathcal{V}$ be the set of variables, and $\mathcal{A}$ be the set of actions. Let  $\mu^i_{k} |v = \mathds{E}_{D_{y}}[y_k \cdot \mathds{1}_{x_i=v}]$ be the expected reward of the action $k$ conditioned to the observation of the value $v$ of the variable $x_i$. Let $\mu^i_{k,v} = P(v).\mu^i_{k} |v = \mathds{E}_{D_{x,y}}[y_k \cdot \mathds{1}_{x_i=v}]$ be the expected reward of the action $k$ and the value $v$ of the binary variable $x_i$. The expected reward when using variable $x_i$ to select the best action is the sum of expected rewards of the best actions for each of its possible values: 
\begin {displaymath}
\mu^i = \sum_{v \in \{0,1\}} P(v).\max_k \mu^i_{k}|v = \sum_{v \in \{0,1\}} \max_k \mu^i_{k,v} 
\end {displaymath}
The optimal variable to be used for selecting the best action is:
$i^*= \arg \max_{i \in \mathcal{V}} \mu^i$.

The algorithm {\sc Variable Selection} chooses the best variable. The $\operatorname {Round-robin}$ function sequentially explores the actions in $\mathcal{A}$ (see Algorithm \ref {VS} line 4).
Each time $t_k$ the reward of the selected action $k$ is unveiled, the estimated expected rewards of the played action $k$ and observed values $\hat {\mu}^i_{k,v}$ and all the estimated rewards of variables $\hat{\mu}^i$ are updated (see VE function lines 2-7). This {\it parallel exploration strategy} allows the algorithm to explore efficiently the variable set. 
When all the actions have been played once (see VE function line 8), irrelevant variables are eliminated if:
\begin {equation}
 \hat{\mu}^{i'}- \hat{\mu}^i +\epsilon \geq 4 \sqrt {\frac {1}{2t_k} \log \frac {4KMt_k^2}{\delta}} \text { ,} 
\label {VE}
\end {equation}
where $i'=  \arg \max_i  \hat{\mu}^i$, and $t_k$ is the number of times the action $k$ has been played.

\begin{algorithm}[h]
   \caption{{\sc Variable Selection}}
   \label{VS}
{\bf Inputs:} $\epsilon \in [0,1)$, $\delta \in (0,1]$\\
{\bf Output:} an $\epsilon$-approximation of the best variable\\
\vspace {-0.5cm}
\begin{algorithmic}[1]
	\STATE $t=0$, $\forall k$ $t_k=0$, $\forall (i,k,v)$ $\hat {\mu}^i_{k,v}=0$, $\forall i$ $\hat{\mu}^i=0$
  \REPEAT
      \STATE {Receive the context vector $\bf x_t$}
      \STATE Play $k=\operatorname{Round-robin}\left(\mathcal{A}\right)$
      \STATE Receive the reward $y_{k}(t)$
      \STATE $t_k=t_k+1$
      \STATE $\mathcal{V}$={VE}$(t_k,k,{\bf x_t},y_k,\mathcal{V},\mathcal{A})$
			\STATE $t=t+1$
  \UNTIL{$|\mathcal{V}|=1$}
	\end{algorithmic}
	\end{algorithm}
	\begin{algorithm}[h]
	\begin {algorithmic} [1]
	\STATE {\bf Function} {VE}$(t,k,{\bf x_t},y_{k},\mathcal{V},\mathcal{A})$
  \FOR {each remaining variable $i \in \mathcal{V}$}
   		\FOR {each value $v$}
   			\STATE $\hat {\mu}^i_{k,v} = \frac {y_{k}}{t}\mathds{1}_{x_i=v}+\frac{t-1}{t}\hat {\mu}^i_{k,v}$
   		\ENDFOR
   		\STATE $\hat{\mu}^i =  \sum_{v \in \{0,1\}} \max_k \hat {\mu}^i_{k,v}$
   \ENDFOR
  \IF { {$k=\operatorname {LastAction}(\mathcal{A}$)}}
      \STATE Remove irrelevant variables from $\mathcal{V}$ according to equation \ref {VE}, or \ref {VE4}
	\ENDIF
	\STATE {\bf return} $\mathcal{V}$
\end{algorithmic}
\end{algorithm}

The parameter $\delta \in (0,1]$ corresponds to the probability of failure. 
The use of the parameter $\epsilon$ comes from practical reasons. The parameter $\epsilon$ is used in order to tune the convergence speed of the algorithm. 
In particular, when two variables provide the highest expected reward, the use of $\epsilon > 0$ ensures that the algorithm stops.
The value of $\epsilon$ has to be in the same order of magnitude as the best mean reward we want to select.
In the analysis of algorithms, we will consider the case where $\epsilon=0$.
Lemma~1 analyzes the sample complexity (the number of iterations before stopping) of {\sc Variable Selection}.

\paragraph {\bf Lemma 1:} {\it when $K\geq 2$, $M \geq 2$, and $\epsilon=0$, the sample complexity of {\sc Variable Selection} needed to obtain $\mathds{P}\left(i' \neq i^*\right) \leq \delta $ is:
\begin {displaymath}
t^* =  \frac {64K}{\Delta_1^2} \log \frac{8KM}{\delta\Delta_1}  \text{, where}\quad \Delta_1 =  \min_{i \neq i^*}(\mu^{i^*} - \mu^{i})
\end {displaymath}
}

Lemma~1 shows that the dependence of the sample complexity needed to select the optimal variable is in $O(\log M)$.
This means that {\sc Variable Selection} can be used to process large set of contextual variables, and hence can be easily extended to categorical variables,
through a binary encoding with only a logarithmic impact on the sample complexity.
Finally, Lemma~2 shows that the variable selection algorithm is optimal up to logarithmic factors.

\paragraph {\bf Lemma 2:} {\it There exists a distribution $D_{x,y}$ such that any algorithm finding the optimal variable $i^*$ has a sample complexity of at least:
\begin {displaymath}
\Omega \left( \frac {K}{\Delta_1^2} \log \frac{1}{\delta}  \right)
\end {displaymath}
}


\subsection {Action selection}

To complete a decision stump, one needs to provide an algorithm which optimizes the choice of the best action knowing the selected variable. 
Any stochastic bandit algorithm such as {\sc UCB} (see \cite {ABF02}), {\sc TS} (see \cite {T33,KKM12}) or {\sc BESA} (see \cite {BMM14}) can be used. For the consistency of the analysis, 
we choose {\sc Successive Elimination} in \cite {EMM02,EDMN06} (see Algorithm~\ref{AS}), that we have renamed {\sc Action Selection}.
Let $\mu_k=\mathds{E}_{D_{y}}[y_k]$ be the expected reward of the action $k$ taken with respect to $D_y$. 
The estimated expected reward of the action is denoted by $\hat{\mu}_k$. 

\begin{algorithm}[h]
   \caption{{\sc Action Selection}}
   \label{AS}
{\bf Inputs:} $\epsilon \in [0,1)$, $\delta \in (0,1]$\\
{\bf Output:} an $\epsilon$-approximation of the best arm\\
\vspace {-0.5cm}
\begin{algorithmic}[1]
  \STATE $t=0$, $\forall k$ $\hat{\mu}_k=0$ and $t_k=0$
  \REPEAT
      \STATE Play $k=\operatorname{Round-robin}\left(A\right)$
      \STATE Receive the reward $y_{k}(t)$ 
      \STATE $t_k=t_k+1$
      \STATE  $\mathcal{A}$={AE}$(t_k,k,y_{k}(t),\mathcal{A})$
			\STATE $t=t+1$
  \UNTIL{$|\mathcal{A}|=1$}
  	\end{algorithmic}
	\end{algorithm}
	\begin{algorithm}[h]
	\begin {algorithmic} [1]
	\STATE {\bf Function} {AE}$(t,k,{\bf x_t},y_{k},\mathcal{A})$
   		\STATE $\hat {\mu}_{k} = \frac {y_{k}}{t}+\frac{t-1}{t}\hat {\mu}_{k}$
      \IF { {$k$={LastAction}($\mathcal{A}$)}}
      		\STATE Remove irrelevant actions from $\mathcal{A}$ according to equation~\ref {AE}, or \ref {AE2}
    	\ENDIF
	\STATE {\bf return} $\mathcal{A}$
\end{algorithmic}
\end{algorithm}

The irrelevant actions in the set $\mathcal{A}$ are successively eliminated when:

\begin {equation}
\hat {\mu}_{k'}- \hat{\mu}_k  +\epsilon \geq 2 \sqrt {\frac {1}{2t_k} \log \frac {4K t_k^2} {\delta}} \text { , }
\label {AE}
\end {equation}
where $k'=\arg \max_k \hat{\mu}_k$, and $t_k$ is the number of times the action $k$ has been played.

\paragraph {\bf Lemma 3:} {\it when  $K\geq 2$, and $\epsilon=0$, the sample complexity of {\sc Action Selection} needed to obtain $\mathds{P}\left( k' \neq k^*\right) \leq \delta$ is:
\begin {displaymath}
t^* =  \frac {64K}{\Delta_2^2} \log \frac{4K}{\delta \Delta_2}  \text{, where}\quad \Delta_2 = \min_{k} (\mu_{k^*} - \mu_k).
\end {displaymath}
}

The proof of Lemma~3 is the same than the one provided for {\sc Successive Elimination} in \cite {EDMN06}.
Finally, Lemma~4 states that the action selection algorithm is optimal up to logarithmic factors (see \cite {MT04} Theorem~1 for the proof).

\paragraph {\bf Lemma 4:} {\it There exists a distribution $D_{x,y}$ such that any algorithm finding the optimal action $k^*$ has a sample complexity of at least:
\begin {displaymath}
\Omega \left( \frac {K}{\Delta_2^2} \log \frac{1}{\delta}  \right)
\end {displaymath}
}
\vspace{-0.7cm}
   
\subsection {Analysis of a decision stump}

The decision stump uses the values of a contextual variable to choose the actions.
The optimal decision stump uses the best variable to choose the best actions. It plays at time $t$:
$k^*_t= \arg \max_k \mu^{i^*}_{k} | v$, where $i^*=\arg \max_i \mu^i$, and $v=x_{i^*}(t)$. The expected gain of the optimal policy is:
\begin {displaymath}
 \mathds{E}_{D_{x,y}} \left[ y_{k^*_t}(t)\right] = \sum_v P(v) \mu^{i^*}_{k^*} | v = \sum_v \mu^{i^*}_{k^*,v} = \mu^{i^*}
\end {displaymath}

To select the best variable, one needs to find the best action of each value of each variable. 
In {\sc Decision Stump} algorithm (see Algorithm \ref {IDS}), an action selection task is allocated for each value of each contextual variable. 
When the reward is revealed, all the estimated rewards of variables $\hat {\mu}^i$ and the estimated rewards of the played action knowing the observed values of variables 
$\hat {\mu}^i_k|v$  are updated (respectively in VE and AE functions): the variables and the actions are simultaneously explored.
However, the elimination of actions becomes effective only when the best variable is selected. 
Indeed, if an action $k$ is eliminated for a value $v_0$ of a variable $i$, the estimation of $\hat {\mu}^i_{k,v_1}$, the mean expected reward of the action $k$ for the value $v_1$, is biased.
As a consequence, if an action is eliminated during the exploration of variables, the estimation of the mean reward $\mu^i$ can be biased. 
That is why, the lower bound of decision stump problem is the sum of lower bound of variable and action selection problems (see Theorem~2).
The only case where an action can be eliminated before the best variable be selected, is when this action is eliminated for all values of all variables. 
For simplicity of the exposition of the {\sc Decision Stump} algorithm we did not handle this case here.

\begin{algorithm}[h]
   \caption{{\sc Decision Stump}}
   \label{IDS}
\begin{algorithmic}[1]
	\STATE $t=0$, $\forall k$ $t_k=0$, $\forall i$ $\hat{\mu}^i=0$, $\forall (i,v,k)$ $t_{i,v,k}=0$, $\mathcal{A}_{i,v}=\mathcal{A}$, $\hat {\mu}^i_{k,v}=0$ and $\hat{\mu}^i_k|v=0$
  \REPEAT
      \STATE {Receive the context vector $\bf x_t$}
      \STATE {\bf if} {$|\mathcal{V}|>1$} {\bf then} Play $k=\operatorname{Round-robin}\left(\mathcal{A}\right)$ 
      \STATE {\bf else}  Play $k=\operatorname{Round-Robin}\left(\mathcal{A}_{i,v}\right)$
      \STATE Receive the reward $y_{k}(t)$
      \STATE $t_k=t_k+1$
  		\STATE $\mathcal{V}$={VE}$\left(t_k,k,{\bf x_t}, y_{k},\mathcal{V},\mathcal{A}\right)$
      \FOR {each variable $i \in \mathcal{V}$}
				\STATE $v=x_i(t)$, $t_{i,v,k}=t_{i,v,k}+1$
      	\STATE $\mathcal{A}_{i,v}=$ {AE}$(t_{i,v,k},k,{\bf x_t}, y_{k},\mathcal{A}_{i,v})$
      \ENDFOR
      \STATE $t=t+1$
  \UNTIL{$t=T$}
\end{algorithmic}
\end{algorithm}

\paragraph {\bf Theorem 1:} {\it when $K\geq 2$, $M \geq 2$, and $\epsilon=0$, the sample complexity  needed by {\sc Decision Stump} to obtain $\mathds{P}\left(i' \neq i^*\right) \leq \delta $ and $\mathds{P}\left( k_t' \neq k_t^*\right) \leq \delta$ is:
\begin {displaymath}
t^* = \frac {64K}{ \Delta_1^2} \log \frac{4KM}{\delta\Delta_1} + \frac {64K}{ \Delta_2^2} \log \frac{4K}{\delta\Delta_2} \text{,}
\end {displaymath}
\begin {center}
where $\Delta_1= \min_{i \neq i^*}\left( \mu^{i^*} - \mu^{i} \right)$,\\
and $\Delta_2 = \min_{k \neq k^*, v \in \{0,1\}} \left(\mu^{i^*}_{k^*,v} - \mu^{i^*}_{k,v}\right)$.
\end {center}
}
Theorem~2 provides a lower bound for the sample complexity, showing that the factors ${1}/{\Delta_1^2}$ and ${1}/{\Delta_2^2}$ are inherent of the decision stump problem. Notice that for the linear bandit problem, the same factor ${1}/{\Delta^2}$ was obtained in the lower bound (see Lemma~2 in \cite {SLM14}).

\paragraph {\bf Theorem 2:} {\it It exists a distribution $D_{x,y}$ such that any algorithm finding the optimal decision stump has a sample complexity of at least:
\begin {displaymath}
\Omega \left( \left( \frac {1}{\Delta_1^2} + \frac {1}{\Delta_2^2}\right) K\log \frac {1}{\delta}  \right)
\end {displaymath}
\begin {center}
\end {center}
}

{\bf Remark 1:}
The factors in $\log 1/\Delta_1$,  $\log 1/\Delta_2$ and $\log 1/\Delta$ could vanish in Theorem~1 following the same approach as that {\it Median Elimination} in \cite {EDMN06}.  Despite the optimality of this algorithm, we did not choose it for the consistency of the analysis.  Indeed, as it suppresses  $1/4$ of variables (or actions) at the end of each elimination phase, {\it Median Elimination} is not well suited when a few number of variables (or actions) provide lot of rewards and the others not. In this case this algorithm spends a lot of times to eliminate non-relevant variables. This case is precisely the one, where we would like to use a local model such as a decision tree.

{\bf Remark 2:}
The extension of the analytical results to an $\epsilon$-approximation of the best decision stump is straightforward using $\Delta_\epsilon=\max (\epsilon,\Delta)$.

\section {\sc Bandit Forest}

A decision stump is a weak learner which uses only one variable to decide the best action. 
When one would like to combine $D$ variables to choose the best action, a tree structure of $\binom {M}{D}.2^D$ multi-armed bandit problems has to be allocated.
To explore and exploit this tree structure with limited memory resources, our approach consists in combining greedy decision trees.
When decision stumps are stacked in a greedy tree, they combine variables in a greedy way to choose the action. When a set of randomized trees vote, they combine variables in a non-greedy way to choose the action.

As highlighted by empirical studies (see for instance \cite {FDCB14}), {\it random forests} of \cite {B01} have emerged as a serious competitors to state-of-the-art methods for classification tasks. 
In \cite {B12}, the analysis of {\it random forests} shows that the reason of these good performances comes from the fact that the convergence of its learning procedure is consistent, and its rate of convergence depends only on the number of strong features, which is assumed to be low in comparison to the number of features.
In this section, we propose to use a {\it random forest} built with the knowledge of $D_{x,y}$ as a reference for the proposed algorithm {\sc Bandit Forest}. 

\begin{algorithm}[h]
   \caption{$\theta$-$\operatorname{OGT}\left( c^*_{\theta},d_\theta, \mathcal{V}_{c^*_\theta}\right)$}
   \label{TOG}
{\bf Inputs: $\theta \in [0,1]$ }\\
{\bf Output:} the $\theta$-optimal greedy tree when $\theta$-$\operatorname{OGT}\left( (),0, \mathcal{V}\right)$ is called\\
\vspace {-0.5cm}
\begin{algorithmic}[1]
\IF {$d_\theta<D_\theta$}
	\STATE $\mathcal{S}_{c^*_\theta}=\left\{ i \in \mathcal{V}_{c^*_\theta} : f(\theta,c^*_\theta,i) = 1 \right\}$
	\STATE $i_{d_\theta+1}^*= \arg \max_{i \in \mathcal{S}_{{c^*_{\theta}}}} \mu^i|c^*_{\theta}$
  \STATE $\theta$-$\operatorname{OGT}\left(c^*_{\theta}+ (x_{i_{d_\theta+1}^*}=0),d_\theta+1, \mathcal{V}_{c^*_\theta} \setminus {\{i_{d_\theta+1}^*\}} \right)$ 		
  \STATE $\theta$-$\operatorname{OGT}\left(c^*_{\theta}+ (x_{i_{d_\theta+1}^*}=1),d_\theta+1,\mathcal{V}_{c^*_\theta} \setminus {\{i_{d_\theta+1}^*\}} \right)$
  \STATE {\bf else }$k^*|c^*_{\theta}= \arg \max_k \mu_k|c^*_{\theta}$
\ENDIF
\end{algorithmic}
\end{algorithm}

Let $\Theta$ be a random variable, which is independent of ${\bf x}$ and ${\bf y}$. Let $\theta \in [0,1]$ the value of $\Theta$. Let $D_\theta$ be the maximal depth of the tree $\theta$.
Let $c_\theta=\{(x_{i_1}=v),...,(x_{i_{d_\theta}}=v)\}$ be the index of a path of the tree $\theta$. We use $c_\theta({\bf x_t})$ to denote the path $c_\theta$ selected at time $t$ by the tree $\theta$. 
Let $f(\theta,i,j): [0,1] \times \{0,1\}^{2^D} \times M \rightarrow \{0,1\}$ be a function which  parametrizes $\mathcal{V}_{c_\theta}$ the sets of available variables for each of $2^{D_\theta}$ splits. 
We call $\theta$-optimal greedy tree, the greedy tree built with the 
knowledge of $D_{x,y}$ and conditioned to the value $\theta$ of the random variable $\Theta$  (see Algorithm \ref {TOG}). 
We call optimal {\it random forest} of size $L$, a {\it random forest}, which consists of a collection of $L$ $\theta$-optimal greedy trees. 
At each time step, the optimal {\it random forest} chooses the action $k^*_t$, which obtains the higher number of votes:
\begin {displaymath}
k^*_t = \arg \max_k \sum_{i=1}^L \mathds{1}_{k^*_{\theta_i,t}=k} \text { ,}
\end {displaymath}
where $k^*_{\theta_i,t}$ is the action chosen by the optimal greedy tree $\theta_i$ at time $t$.

\begin{algorithm}[h]
   \caption{{\sc Bandit Forest}}
   \label{Bandit Forest}
\begin{algorithmic}[1] 
   \STATE $t=0$, $\forall \theta$ $c_{\theta}=()$ and $d_\theta=0$, $\forall \theta$ {NewPath}$(c_{\theta},\mathcal{V})$
   \REPEAT
      \STATE Receive the context vector $\bf x_t$
			\STATE {\bf for} {each $\theta$} {\bf do} select the context path $c_{\theta}({\bf x_t})$
			\STATE {\bf if} {$\forall \theta$ $d_\theta = D_{\theta}$ and $|\mathcal{S}_{c_{\theta}}|=1$, and $\forall (\theta,i,v)$ $|\mathcal{A}_{c_{\theta,i,v}}|=1$}
						 {\bf then} $k = \arg \max_k \sum_{i=1}^L \mathds{1}_{k_{\theta_i,t}=k}$
			\STATE {\bf else} $k=\operatorname{Round-robin}(\mathcal{A})$
			\STATE {\bf endif}
			\STATE Receive the reward $y_k(t)$
			\FOR {each $\theta$}
								\STATE {$t_{c_{\theta},k}=t_{c_{\theta},k}+1$}
								\STATE $\mathcal{S}_{c_\theta}$={VE}$\left(t_{c_{\theta},k},k,{\bf x_t}, y_{k},\mathcal{S}_{c_\theta},\mathcal{A}\right)$
								\FOR {each remaining variable $i$}
												\STATE $v=x_i(t)$, $t_{c_\theta,i,v,k}=t_{c_\theta,i,v,k}+1$
      									\STATE $\mathcal{A}_{c_\theta,i,v}=$ {AE}$(t_{c_\theta,i,v,k},k,{\bf x_t}, y_{k},\mathcal{A}_{c_\theta,i,v})$
								\ENDFOR
								\IF {$|\mathcal{S}_{c_{\theta}}|=1$ \AND $d_\theta < D_{\theta}$}
												\STATE $\mathcal{V}_{c_\theta}=\mathcal{V}_{c_\theta} \setminus \{i\}$
												\STATE {NewPath}$(c_{\theta} + \left(x_{i}=0\right),\mathcal{V}_{c_\theta})$
												\STATE {NewPath}$(c_{\theta} + \left(x_{i}=1\right),\mathcal{V}_{c_\theta})$	
								\ENDIF
			\ENDFOR
	 \STATE $t=t+1$
   \UNTIL {$t = T$}
\end{algorithmic}
\end{algorithm}
\begin{algorithm}[h]
	\begin {algorithmic} [1]
	\STATE {\bf Function} {NewPath}$(c_\theta,\mathcal{V}_{c_\theta})$
	 \STATE $\mathcal{S}_{c_\theta}=\left\{ i \in \mathcal{V}_{c_\theta} : f(\theta,c_\theta,i) = 1 \right\}$
	 \STATE $d_\theta=d_\theta+1$, $\forall (i,v)$ $\mathcal{A}_{c_\theta,i,v}=\mathcal{A}$
	 \STATE $\forall k$ $t_{c_\theta,k}=0$, $\forall (i,v,k)$ $t_{c_\theta,i,v,k}=0$, $\forall i$ $\hat{\mu}^i|c_\theta=0$
   \STATE $\forall (i,k,v)$ $\hat {\mu}^i_{k,v}|c_\theta=0$ and $\hat{\mu}^i_k|(v,c_\theta)=0$
\end{algorithmic}
\end{algorithm}

{\sc Bandit Forest} algorithm explores and exploits a set of $L$ decision trees knowing $\theta_1,...,\theta_L$ (see Algorithm \ref {Bandit Forest}).
When a context ${\bf x_t}$ is received  (line 3):
\begin {itemize}
\item For each tree $\theta$, the path $c_\theta$ is selected (line 4).
\item An action $k$ is selected:

\begin {itemize}
\item  If the learning of all paths $c_\theta$ is finished, then each path vote for its best action  (line 5).
\item  Else the actions are sequentially played (line 6).
\end {itemize}

\item The reward $y_k(t)$ is received  (line 8).
\item The decision stumps of each path $c_\theta$ are updated (lines 9-15).
\item When the set of remaining variables of the decision stump corresponding to the path $c_\theta$ contains only one variable and the maximum depth $D_\theta$ is not reached (line 16), two new decision stumps corresponding to the values {0,1} of the selected variable are allocated (lines 18-20). The random set of remaining variables $\mathcal{V}_{c_\theta}$, the counts, and the estimated means are initialized in function NewPath. 
\end {itemize}

To take into account the $L$ decision trees of maximum depth $D_\theta$, irrelevant variables are eliminated using a slight modification of inequality (\ref {VE}).
A possible next variable $x_i$ is eliminated when:

\begin {equation}
 \hat{\mu}^{i'}|c_{\theta}- \hat{\mu}^i|c_{\theta} + \epsilon \geq 4 \sqrt {\frac {1}{2t_{c_{\theta},k}} \log \frac {4\times 2^DKMD_\theta Lt^2_{c_{\theta},k}}{\delta}} \text { ,}
\label {VE4}
\end {equation}
where $i'=  \arg \max_{i \in \mathcal{V}_{c_{\theta}}} \hat{\mu}^i|c_{\theta}$, and $t_{c_{\theta},k}$ is the number of times the path $c_{\theta}$ and the action $k$ have been observed.
To take into account the $L$ decision trees, irrelevant actions are eliminated using a slight modification of inequality (\ref {AE}):

\begin {equation}
\hat {\mu}_{k'|c_{\theta}}- \hat{\mu}_{k|c_{\theta}}  +\epsilon \geq 2 \sqrt {\frac {1}{2t_{c_{\theta},i,v,k}} \log \frac {4\times 2^DKL t_{c_{\theta},i,v,k}^2} {\delta}} \text { ,}
\label {AE2}
\end {equation}
where $k'=\arg \max_k \hat{\mu}_{k|c_{\theta}}$, and $t_{c_{\theta},i,v,k}$ is the number of times  the action $k$ has been played when the path $c_{\theta}$, and the value $v$ of the variable $i$ have been observed.

\paragraph {\bf Theorem 3:} {\it when $K\geq 2$, $M \geq 2$, and $\epsilon=0$, the sample complexity needed by {\sc Bandit Forest} learning to obtain the optimal {\it random forest} of size $L$ with a probability at least $1-\delta$ is:
\begin {displaymath}
t^* =   2^D \left(\frac {64K}{ \Delta_1^2} \log \frac{4KMDL}{\delta \Delta_1} + \frac {64K}{ \Delta_2^2} \log \frac{4LK}{\delta\Delta_2}\right) \text{ ,}
\end {displaymath}
\begin {center}
where $\Delta_{1}=\min_{\theta,c^*_\theta,i \neq i^*}P(c^*_\theta)\left( \mu^{i^*} | c^*_{\theta} - \mu^{i} | c^*_{\theta} \right)$, $\Delta_{2} = \min_{\theta,c^*_\theta,k \neq k^*} P(c^*_\theta)\left( \mu_{k^*} |c^*_{\theta} - \mu_{k} | c^*_{\theta}\right)$, and $D=\max_{\theta} D_{\theta}$.
\end {center}
}

The dependence of the sample complexity on the depth $D$ is exponential. 
This means that like all decision trees, {\sc Bandit Forest} is well suited for cases,  where there is a small subset of relevant variables belonging to a large set of variables ($D<<M$). 
This usual restriction of local models is not a problem for a lot of applications, where one can build thousands of contextual variables, and where only a few of them are relevant. 

\paragraph {\bf Theorem 4:}
{\it  There exists a distribution $D_{x,y}$ such that any algorithm finding the optimal {\it random forest} of size $L$ has a sample complexity of at least:
\begin {displaymath}
\Omega\left( 2^D \left[ \frac {1}{\Delta_1^2} + \frac {1}{\Delta_2^2}\right] K\log \frac{1}{\delta} \right)
\end {displaymath}
\begin {center}
\end {center}
}
\vspace{-0.3cm}
Theorem~4 shows that {\sc Bandit Forest} algorithm is near optimal. 
The result of this analysis is supported by empirical evidence in the next section.

{\bf Remark 4:} We have chosen to analyze {\sc Bandit Forest} algorithm in the case of $\epsilon=0$ in order to simplify the concept of the optimal policy. Another way is to define the set of $\epsilon$-optimal {\it random forests}, built with decision stumps which are optimal up to an $\epsilon$ approximation factor. In this case, the guarantees are given with respect to the worst policy of the set. When $\epsilon=0$, this set contains only the optimal {\it random forest} of size $L$ given the values of $\Theta$.

\section {Experimentation}

In order to illustrate the theoretical analysis with reproducible results on large sets of contextual variables,
we used three datasets from the {\it UCI Machine Learning Repository} ({\it Forest Cover Type}, {\it Adult}, and {\it Census1990}).
We recoded each continuous variable using {\it equal frequencies} into $5$ binary variables, and each categorical variable into disjunctive binary variables.
We obtained $94$, $82$ and $255$ binary variables, for {\it Forest Cover Type}, {\it Adult} and {\it Census1990} datasets respectively.
For {\it Forest Cover Type}, we used the $7$ target classes as the set of actions. For {\it Adult}, the categorical variable {\it occupation} is used as a set of $14$ actions.
For {\it Census1990}, the categorical variable {\it Yearsch} is used as a set of $18$ actions.
The gain of policies was evaluated using the class labels of the dataset with a reward of $1$ when the chosen action corresponds to the class label and $0$ otherwise.
The datasets, respectively composed of $581000$, $48840$ and $2458285$ instances, were shuffled and played in a loop to simulate streams. 
In order to introduce noise between loops, at each time step the value of each binary variable has a probability of $0.05$ to be inverted. 
Hence, we can consider that each context-reward vector is generated by a stationary random process.
We set the time horizon to $10$ millions of iterations. The algorithms are assessed in terms of accumulated regret against the optimal {\it random forest} of size $100$. 
The optimal {\it random forest} is obtained by training a {\it random forest} of size $100$ not limited by depth on the whole dataset with full information feedback and without noise.

\begin{figure}[h]
\centering
   \includegraphics[width=7.6cm]{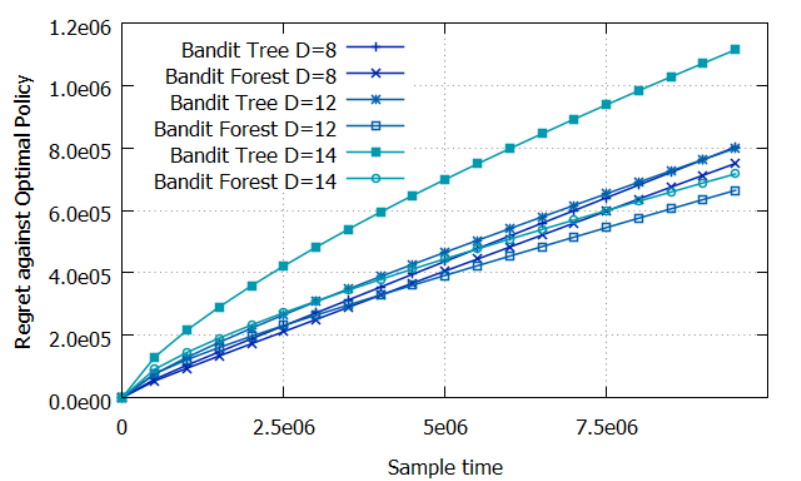}
   \vspace{-0.1cm}
\caption{The accumulated regret of  {\sc Bandit Forest} and {\sc Bandit Tree} with different depths 
against the optimal policy averaged over ten trials on {\it Forest Cover Type} dataset played in a loop with noisy inputs.}
   \label{Cov3}
\end {figure}

\begin{figure}[h]
\centering
   \includegraphics[width=7.6cm]{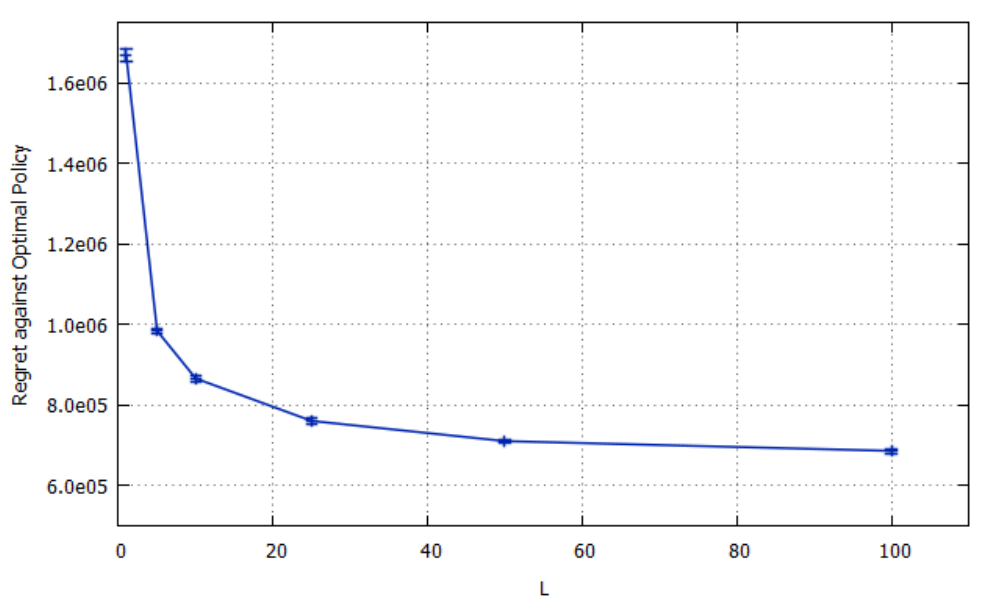}
      \vspace{-0.1cm}
\caption{Sensitivity to the size of {\sc Bandit Forest} on {\it Census 1990} dataset.}
   \label{Census2}
\end {figure}

To effectively implement {\sc Bandit Forest}, we have done two modifications of the analyzed algorithms. 
Firstly, the Round-robin function is replaced by an uniform random draw from the union of the remaining actions of the paths.
Secondly, the rewards are normalized using {\it Inverse Propensity Scoring} (see \cite {HT52}): the obtained reward is divided by the probability to draw the played action.
First of all, notice that the regret curves of {\sc Bandit Forest} algorithm are far from those of explore then exploit approaches: the regret is gradually reduced over time (see Figure \ref {Cov3}). Indeed, {\sc Bandit Forest} algorithm uses a {\it localized explore then exploit} approach: most of paths, which are unlikely, may remain in exploration state, while the most frequent ones already vote for their best actions. The behavior of the algorithm with regard to number of trees $L$ is simple: the higher $L$, the greater the performances, and the higher the computational cost (see Figure \ref {Census2}).
To analyze the sensitivity to depth of {\sc Bandit Forest} algorithm, we set the maximum depth and we compared the performances of a single tree without randomization ({\sc Bandit Tree}) and of {\sc Bandit Forest} with $L=100$. The trees of the forest are randomized at each node with different values of $\epsilon$ (between $0.4$ and 0.8), and with different sets of available splitting variables (random subset of $80 \%$ of remaining variables). For each tested depth, a significant improvement is observed thanks to the vote of randomized trees (see Figure \ref {Cov3}).  Moreover, {\sc Bandit Forest} algorithm appears to be less sensible to depth than a single tree: the higher the difference in depth, the higher the difference in performance.

\begin{figure}[h]
\centering
   \includegraphics[width=7.6cm]{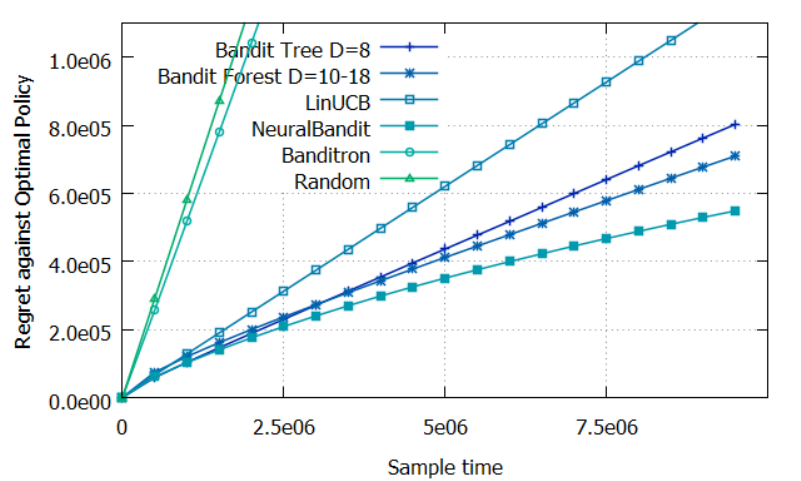}
   \vspace{-0.1cm}
\caption{The accumulated regret against the optimal policy averaged over ten trials for the dataset {\it Forest Cover Type} played in a loop with noisy inputs.}
   \label{Cov2}
\end {figure}

\begin{figure}[h]
\centering
   \includegraphics[width=7.6cm] {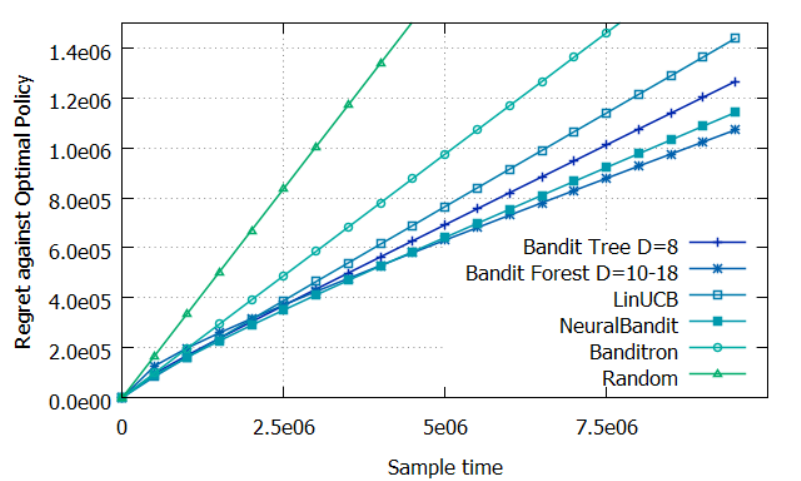}
   \vspace{-0.1cm}
\caption{The accumulated regret against the optimal policy averaged over ten trials for the dataset {\it Adult} played in a loop with noisy inputs.}
   \label{Adult}
\end{figure}

\begin{figure}[h]
\centering
   \includegraphics[width=7.6cm] {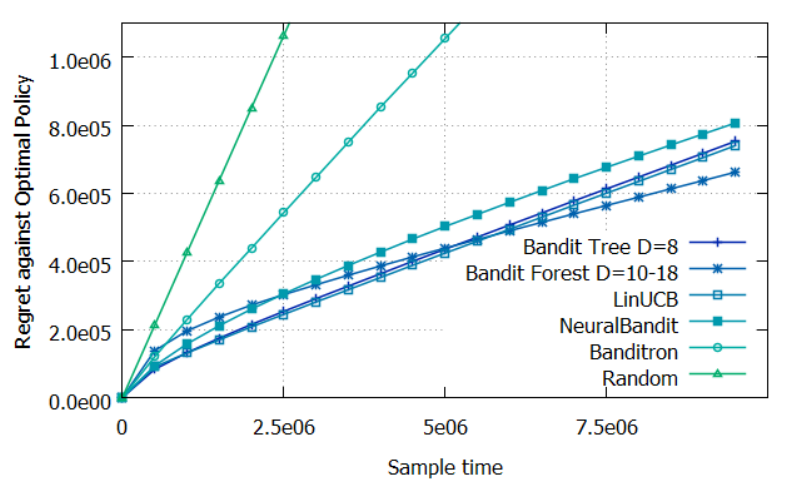}
   \vspace{-0.1cm}
\caption{The accumulated regret against the optimal policy averaged over ten trials for the dataset {\it Census1990} played in a loop with noisy inputs.}
   \label{Census}
\end{figure}

To compare with state-of-the-art, the trees in the forest are randomized with different values of the parameters $D$ (between $10$ and $18$). 
On the three datasets, {\sc Banditron} (\cite {{KST08}}) is clearly outperformed by the other tested algorithms (see Figures \ref {Cov2}, \ref {Adult}, \ref {Census}). 
{\sc Neural Bandit} (\cite {AFB14}) is a Committee of Multi-Layer Perceptrons (MLP). Due to the non convexity of the error function, MLP trained with
the back-propagation algorithm (\cite {RHW86}) does not find the optimal solution. That is why finding the regret bound of such model is still an open problem. However,
since MLPs are universal approximators (\cite {HSW89}), {\sc Neural Bandit} is a very strong baseline. It outperforms {\sc LinUCB} (\cite {LCLS10}) on {\it Forest Cover Type} and on {\it Adult}.
{\sc Bandit Forest} clearly outperforms {\sc LinUCB} and {\sc Banditron} on the three datasets.
In comparison to {\sc Neural Bandit}, {\sc Bandit Forest} obtains better results on {\it Census 1990} and {\it Adult}, and it is outperformed on {\it Forest Cover Type}, where the number of strong features seems to be high. 
Finally as shown by the worst case analysis, the risk to use {\sc Bandit Forest} on a lot of optimization problems is controlled, 
while {\sc Neural Bandit}, which has no theoretical guaranty, can obtain poor performances on some problems, such as {\it Census 1990} where it is outperformed by the linear solution {\sc LinUCB}. 
This uncontrolled risk increases with the number of actions,  since the probability to obtain a non robust {\sc MLP} linearly increases with the number of actions.

\section {Conclusion}

We have shown that the proposed algorithm is optimal up to logarithmic factors with respect to a strong reference: a {\it random forest} built knowing the joint distribution of the contexts and rewards. 
In the experiments, {\sc Bandit Forest} clearly outperforms {\sc LinUCB}, which is a strong baseline with known regret bound, 
and performs as well as {\sc Neural Bandit}, for which we do not have theoretical guaranty.
Finally, for applications where the number of strong features is low in comparison to the number of possible features, which is often the case, 
{\sc Bandit Forest} shows valuable properties: 
      \vspace{-0.2cm}
\begin {itemize}
\item its sample complexities have a logarithmic dependence on the number of contextual variables, which means that
it can process a large amount of contextual variables with a low impact on regret, 
\item its low computational cost allows to process efficiently infinite data streams, 
\item like all decision tree algorithms, it is well suited to deal with non linear dependencies between contexts and rewards.
\end {itemize}

\bibliographystyle{spmpsci}  

\onecolumn
\section {Appendix}

\subsection {Notations}

In order to facilitate the reading of the paper, we provide below (see Table \ref{tab2}) the list of notations.

\begin {center}
\begin{table} [htb]
\caption{Notations}
\label{tab2}  
\begin{tabular}{lll}
\hline\noalign{\smallskip}
notation & description\\
\noalign{\smallskip}
\hline
\noalign{\smallskip}
$K$ & number of actions \\
$M$ & number of contextual variables \\
$D_\theta$ & maximum depth of the tree $\theta$\\
$L$ & number of trees \\
$T$ & time horizon \\
$\mathcal{A}$ & set of actions\\
$\mathcal{V}$ & set of variables\\
$\mathcal{S}$ & set of remaining variables \\
$\bf x$ & context vector ${\bf x}=(x_1,\ldots,x_M)$\\
$\bf y$& reward vector ${\bf y}=(y_1,\ldots,y_K)$\\
$k_t$ & action chosen at time $t$\\
$c_\theta$ & context path the tree $\theta$, $c_\theta=(x_{i_1},v_{i_1}),...,(x_{i_{d_\theta}},v_{i_{d_\theta}})$ \\
$d_\theta$ & current depth of the context path $c_\theta$\\
$\mu_k$ & expected reward of action $k$, $\mu_k= \mathds{E}_{D_y}[y_k]$\\
$\mu^i_{k}|v$  & expected reward of action $k$ conditioned to $x_i=v$, $\mu^i_{k}|v= \mathds{E}_{D_{y}}[y_k \cdot \mathds{1}_{x_i=v}]$\\
$\mu^i_{k,v}$  & expected reward of action $k$ and $x_i=v$, $\mu^i_{k,v}= \mathds{E}_{D_{x,y}}[y_k \cdot \mathds{1}_{x_i=v}]$\\
$\mu^i$  & expected reward for the use of the variable $x_i$ to select the best actions\\
$\delta$ & probability of error \\
$\epsilon $ & approximation error \\
$\Delta_1$ & minimum of difference with the expected reward of the best action $\mu_{k^*}$ \\
& and the expected reward of a given action $k$: $\Delta_1 =  \min_{k \neq k^*}(\mu_{k^*} - \mu_k$)\\
$\Delta_2$ & minimum of difference with the best variable expected reward $\mu^{i^*}$\\
&  and the expected reward for other variables: $\Delta_2 =  \min_{i \neq i^*}(\mu^{i^*} - \mu^i)$\\
$t^*$ & sample complexity of the decision stump\\
\hline
\end{tabular}
\end{table}
\end {center}

\subsection {Lemma~1}

\begin {proof}

We cannot use directly Hoeffding inequality (see \cite {H63}) to bound the estimated gains of the use of variables.
The proof of Lemma 1 overcomes this difficulty by bounding each estimated gain $\mu^i$ by the sum of the bounds over the values of the expected reward of the best action $\mu^i_{k^*,v}$  (inequality \ref {eq3t1}). From Hoeffding's inequality, at time $t$ we have:

\begin {displaymath}
\begin {split}
 \mathds{P} \left(\left| \hat{\mu}^i_{k,v}  - \mu^i_{k,v } \right| \geq \alpha_{t_k} \right)  \leq 2\exp(-2\alpha_{t_k}^2 t_k) 
 & = \frac {\delta}{4KM t_k^2} \text{ ,}
\end {split}
\end {displaymath}

where $\alpha_{t_k} = \sqrt {\frac {1}{2t_k} \log \frac {4KMt_k^2}{\delta}}$.


Using Hoeffding's inequality on each time $t_k$, applying the union bound and then $\sum 1/t_k^2 = \pi^2/6$, the following inequality holds for any time $t$ with a probability $1-\frac {\delta\pi^2}{12KM}$:

\begin {equation}
\hat{\mu}^i_{k,v}  - \alpha_{t_k} \leq \mu^i_{k,v} \leq \hat{\mu}^i_{k,v}  + \alpha_{t_k}
\label {eq1t1}
\end {equation}

If the inequality (\ref {eq1t1}) holds for the actions $k'=\arg \max_k \hat{\mu}^i_{k,v} $, and $k^*=\arg \max_k \mu^i_{k,v}$, we have:

\begin {equation}
\begin {split}
&   \hat{\mu}^i_{k',v} - \alpha_{t_k} \leq \mu^i_{k',v} \leq \mu^i_{k^*,v}\leq \hat{\mu}^i_{k^*,v} + \alpha_{t_k} \leq \hat{\mu}^i_{k',v}  + \alpha_{t_k} \\
& \Rightarrow  \hat{\mu}^i_{k',v}  - \alpha_{t_k} \leq  \mu^i_{k^*,v} \leq \hat{\mu}^i_{k',v}  + \alpha_{t_k}
\end {split}
\label {eq2t1}
\end {equation}

If the previous inequality (\ref {eq2t1}) holds for all values $v$ of the variable $x_i$, we have:

\begin {equation}
\begin {split}
& \sum_{v \in \{0,1\}} \left( \hat{\mu}^i_{k',v} - \alpha_{t_k} \right) \leq \sum_{v \in \{0,1\}}  \mu^i_{k^*,v} \leq \sum_{v \in \{0,1\}} \left( \hat{\mu}^i_{k',v} + \alpha_{t_k} \right) \\
& \Leftrightarrow \hat{\mu}^i  - 2\alpha_{t_k} \leq \mu^i \leq \hat{\mu}^i  + 2\alpha_{t_k}
\end {split}
\label {eq3t1}
\end {equation}
If the previous inequality holds for $i'=\arg \max_i \hat{\mu}^i$, we have:

\begin {equation}
\hat{\mu}^{i'}  - 2\alpha_{t_k} \leq \mu^{i'}\leq \mu^{i^*}
\label {eq4t1}
\end {equation}

As a consequence, the variable $x_i$ cannot be the best one when:
\begin {equation}
\hat{\mu}^i  + 2 \alpha_{t_k} \leq \hat{\mu}^{i'}  - 2\alpha_{t_k} \text { ,}
\label {eq5t1}
\end {equation}

Using the union bound, the probability of making an error about the selection on the next variable by eliminating each variable $x_i$ when the inequality (\ref {eq5t1}) holds is bounded by the sum for each variable $x_i$ and each value $v$ that the inequality \ref {eq2t1} does not hold for $k'$ and $k^*$:
\begin {equation}
\mathds{P}\left( i^* \neq i' \right) \leq  \sum_{i \in \mathcal{V}} \frac {K\delta\pi^2}{24KM} 
\leq \sum_{i \in \mathcal{V}}  \frac {\delta}{M} \leq \delta
\label {eq2}
\end {equation}

Now, we have to consider $t^*_k$, the number of steps needed to select the optimal variable.
If the best variable has not been eliminated (probability $1-\delta$), the last variable $x_i$ is eliminated when:
\begin {displaymath}
\hat{\mu}^{i^*} -\hat{\mu}^i \geq 4 \alpha_{t^*_k}
\end {displaymath}

The difference between the expected reward of a variable $x_i$ and the best next variable is defined by:

\begin {displaymath}
\Delta_i =  \mu^{i^*} - \mu^i
\end {displaymath}

Assume that:
\begin {equation}
\Delta_i \geq 4 \alpha_{t_k}
\label {eq3}
\end {equation}

The following inequality holds for the variable $x_i$ with a probability $1-\frac {\delta}{KM}$:
\begin {displaymath}
\hat{\mu}^i- 2 \alpha_ {t_k} \leq \mu^i  \leq \hat{\mu}^i + 2 \alpha_{t_k}
\end {displaymath}
Then, using the previous inequality in the inequality (\ref {eq3}), we obtain:

\begin {displaymath}
(\hat{\mu}^{i^*} +  2\alpha_{t_k}) -  (\hat{\mu}^i + 2 \alpha_{t_k}) \geq \mu^{i^*} - \mu^i \geq 4 \alpha_{t_k}
\end {displaymath}

Hence, we have:
\begin {displaymath}
\hat{\mu}^{i^*} -  \hat{\mu}^i  \geq 4 \alpha_{t_k}
\end {displaymath}

The condition $\Delta_i \geq 4 \alpha_{t_k}$ implies the elimination of the variable $x_i$.
Then, we have:
\begin {displaymath}
\Delta_i  \geq 4  \alpha_{t_k}
\end {displaymath}

\begin {equation}
\Rightarrow \Delta^2_i \geq \frac {8}{t_k} \log \frac {4KMt^2_k}{\delta}
\label {eq4}
\end {equation}

The time $t^*_k$, where all non optimal variables have been eliminated, is reached when the variable corresponding to the minimum of $\Delta^2_i$ is eliminated.

\begin {equation}
\Rightarrow \Delta^2 \geq \frac {8}{t^*_k} \log \frac {4KM {t^*_k}^2}{\delta} \text { ,}
\label {eq4b}
\end {equation}
where $\Delta=\min_{i \neq i^*} \Delta_i$.

The inequality (\ref {eq4b}) holds for all variables $i$ with a probability $1-\delta$ for:
\begin {equation}
t^*_k= \frac {64}{\Delta^2} \log \frac {4KM}{\delta.\Delta}
\label {eq5}
\end {equation}

Indeed, if we replace the value of $t^*_k$ in the right term of the inequality (\ref {eq4}), we obtain:

\begin {displaymath}
\begin {split}
& \frac {\Delta^2} {8\log \frac {4KM}{\delta.\Delta}}\left(  \log \frac {4KM}{\delta} + 2 \log \frac {64}{\Delta^2} + 2\log\log \frac {4KM}{\delta.\Delta} \right) = \\
& \frac {\Delta^2} {8\log \frac {4KM}{\delta.\Delta}}\left(  \log \frac {4KM}{\delta} -4 \log \Delta+ 12\log 2  + 2\log\log \frac {4KM}{\delta.\Delta} \right) \leq \\
& \frac {\Delta^2} {8\log \frac {4KM}{\delta.\Delta}}\left(  4 \log \frac {4KM}{\delta.\Delta} + 12\log 2  + 2\log\log \frac {4KM}{\delta.\Delta} \right)
\end {split}
\end {displaymath}

For $x \geq 13$, we have: 
\begin {displaymath}
12\log 2 + 2 \log \log x < 4\log x
\end {displaymath}

Hence, for $4KM \geq 13$, we have:
 
\begin {displaymath}
\begin {split}
& \frac {\Delta^2} {8\log \frac {4KM}{\delta.\Delta}}\left(  4 \log \frac {4KM}{\delta.\Delta} + 12\log 2  + 2\log\log \frac {4KM}{\delta.\Delta} \right) \leq \\
&  \frac {\Delta^2} {8\log \frac {4KM}{\delta.\Delta}} 8 \log \frac {4KM}{\delta.\Delta} = \Delta^2
\end {split}
\end {displaymath}

Hence,  we obtain:
\begin {displaymath}
t^*_k = \frac {64}{ \Delta^2} \log \frac{4KM}{\delta } \text { , with a probability $1-\delta$}
\end {displaymath}

As the actions are chosen the same number of times (Round-robin function), we have $t=Kt_k$, and thus:

\begin {displaymath}
t^* = \frac {64K}{ \Delta^2} \log \frac{4KM}{\delta } \text { , with a probability $1-\delta$}
\end {displaymath}

\end {proof}

\subsection {Lemma~2}

\begin {proof}

Let $y^i_{k,v}$ be a bounded random variable corresponding to the reward of the action $k$ when the value $v$ of the variable $i$ is observed. 
Let $y^i$ be a random variable such that:
\begin {displaymath}
y^i = \max_k y^i_{k,v}
\end {displaymath}

We have:
\begin {displaymath}
\mathds{E}_{D_{x,y}} [y^i] = \mu^i
\end {displaymath}

Each $y^i$ is updated each step $t_k$ when each action has been played once.
Let $\Theta$ be the sum of the binary random variables $\theta_1,...,\theta_{t_k},...,\theta_{t_k^*}$ such that $\theta_{t_k} =  \mathds{1}_{y^i(t_k) \geq y^j(t_k)}$.
Let $p_{ij}$ be the probability that the use of variable $i$ leads to more rewards than the use of variable $j$. We have:

\begin {displaymath}
p_{ij}=\frac {1}{2} - \Delta_{ij} \text { , where $\Delta_{ij} = \mu^i - \mu^j$.}
\end {displaymath}

Slud's inequality (see \cite {S77}) states that when $p\leq 1/2$ and $t_k^* \leq x \leq t_k^*.(1-p)$, we have:
\begin {equation}
P(\Theta \geq x) \geq P\left(Z \geq \frac {x-t_k^*.p}{\sqrt {t_k^*.p(1-p)}} \right) \text { ,} 
\label {Slud1}
\end {equation}
where $Z$ is a normal $\mathcal{N}(0,1)$ random variable.

To choose the best variable between $i$ and $j$, one needs to find the time $t^*_k$ where $P(\Theta \geq t^*_k/2)\geq \delta$.
To state the number of trials $t_k^*$ needed to estimate $\Delta_{ij}$, we recall and adapt the arguments developed in \cite {M10}.
Using Slud's inequality (see equation \ref {Slud1}), we have:

\begin {equation}
P(\Theta \geq t_k^*/2 )\geq P\left(Z \geq \frac {t_k^*.\Delta_{ij}}{\sqrt {t_k^*.p_{ij}(1-p_{ij})}} \right) \text { ,} 
\label {Slud2}
\end {equation}

Then, we use the lower bound of the error function (see \cite {C55}):

\begin {displaymath}
P(Z \geq z) \geq 1-\sqrt{1-\exp\left({-\frac{z^2}{2}}\right)}
\end {displaymath}

Therefore, we have:

\begin {displaymath}
\begin {split}
P(\Theta \geq t_k^*/2 ) & \geq  1-\sqrt{ 1-\exp\left({-\frac {t_k^*.\Delta^2_{ij}}{2p_{ij}(1-p_{ij})}}\right)} \\
& \geq 1-\sqrt{ 1-\exp\left({-\frac {t_k^*.\Delta^2_{ij}}{p_{ij}}}\right)} \\
& \geq \frac {1}{2}\exp\left({-\frac {t_k^*.\Delta^2_{ij}}{p_{ij}}}\right) 
\end {split}
\end {displaymath}

As $p_{ij}=1/2 - \Delta_{ij}$, we have:

\begin {displaymath}
\log \delta = \log \frac {1}{2} -\frac {t_k^*.\Delta^2_{ij}}{1/2-\Delta_{ij}} \geq \log \frac {1}{2} -2t_k^*.\Delta^2_{ij}
\end {displaymath}

Hence, we have:

\begin {displaymath}
t_k^* =  \Omega \left( \frac {1}{\Delta_{ij}^2} \log \frac{1}{\delta}  \right)
\end {displaymath}

Then, we need to use the fact that as all the values of all the variables are observed when one action is played: the $M(M-1)/2$ estimations of bias are solved in parallel.  In worst case, $\min_{ij} \Delta_{ij}=\min_j \Delta_{i^*j}=\Delta$.
Thus any algorithm needs at least a sample complexity $t^*$, where:
\begin {displaymath}
t^* = K.t_k^* =  \Omega \left( \frac {K}{\Delta^2} \log \frac{1}{\delta}  \right)
\end {displaymath}

\end {proof}

\subsection {Theorem~1}

\begin {proof}

Lemma~1 states that the sample complexity needed to find the best variable is:
\begin {displaymath}
t^*_1 =  \frac {64K}{\Delta_1^2} \log \frac{4KM}{\delta\Delta_1}  \text{, where}\quad \Delta_1 =  \min_{i \neq i^*}(\mu^{i^*} - \mu^{i})
\end {displaymath}

Lemma~3 states that the sample complexity needed to find the optimal action for a value $v$ of the best variable is:
\begin {displaymath}
t^*_{2,v} =  \frac {64K}{\Delta_{2,v}^2} \log \frac{4K}{\delta \Delta_{2,v}}  \text{, where}\quad \Delta_{2,v} = \min_{k \neq k^*} (\mu^{i^*}_{k^*,v} - \mu^{i^*}_{k,v}).
\end {displaymath}

The sample complexity of decision stump algorithm is bounded by the sum of the sample complexities of variable selection and action elimination algorithms:

\begin {displaymath}
t^*=t^*_1 + t^*_{2}  \text { , where $t^*_{2}=\max_v t^*_{2,v}$.}
\end {displaymath}

\end {proof}

\subsection {Theorem~2}

\begin {proof}

In worst case, all the values of variables have different best actions, and $K=2M$.
If an action is suppressed before the best variable is selected, the estimation of the mean reward of one variable is underestimated.
In worst case this variable is the best one, and a sub-optimal variable is selected.
Thus, the best variable has to be selected before an action be eliminated.
The lower bound of the decision stump problem is the sum of variable selection and best arm identification lower bounds, stated respectively in Lemma~2 and Lemma~4.

\end {proof}

\subsection {Theorem~3}

\begin {proof}

The proof of Theorem~3 uses Lemma~1 and Lemma~3. Using the slight modification of the variable elimination inequality proposed in section 3,
Lemma~1 states that for each decision stump, we have:

\begin {displaymath}
\mathds{P}\left( i^*_{d_\theta} | c_{\theta} \neq i'_{d_\theta} | c_{\theta}\right) \leq \frac {\delta}{2^{D_\theta} L} 
\end {displaymath} 

For the action corresponding to the path $c_\theta$, Lemma~3 states that:

\begin {displaymath}
\mathds{P}(k \neq k^*) \leq \frac {\delta}{2^{D_\theta} L}
\end {displaymath}
From the union bound, we have:
\begin {displaymath}
\mathds{P}(\exists c_\theta \text{such that} c_\theta \neq c_\theta^*) \leq \delta \text { and } \mathds{P}(\exists k \text{such that} k \neq k^*) \leq \delta
\end {displaymath}
Using Lemma~1 and Lemma~3, and summing the sample complexity of each $2^{D_\theta}$ variable selection tasks and the sample complexity of each $2^{D_\theta}$  action selection tasks,
we bound the sample complexity of any tree $\theta$ by:
\begin {displaymath}
t^* \leq 2^D \frac {64K}{ \Delta_1^2} \log \frac{4KMDL}{\delta\Delta_1} + 2^D \frac {64K}{ \Delta_2^2} \log \frac{4KL}{\delta\Delta_2} \text { ,}
\end {displaymath}
where $D=\max D_\theta$.

\end {proof}

\subsection {Theorem~4}
\begin {proof}

To build a decision tree of depth $D_\theta$, any greedy algorithm needs to solve $\sum_{d<D_\theta} 2^d = 2^{D_\theta}$ variable selection problems (one per node), and $2^{D_\theta}$ action selection problems (one per leaf).
Then, using Lemma~2 and Lemma~4, any greedy algorithm needs a sample complexity of at least:
\begin {displaymath}
t^* \geq \Omega\left( 2^D \left[ \frac {1}{\Delta_1^2} + \frac {1}{\Delta_2^2}\right] K\log \frac {1}{\delta} \right)
\end {displaymath}

\end {proof}

\subsection {Additional experimental results}

We provide below (see Table \ref {tab}) the classification rates to compare the asymptotical performances of each algorithm, and the processing times.

\begin {center}
\begin {table} [hb]
\caption {Summary of results on the datasets played in a loop. The regret against the optimal {\it random forest} is evaluated on ten trials. 
Each trial corresponds to a random starting point in the dataset.
The confidence interval is given with a probability $95 \%$. 
The classification rate is evaluated on the last $100000$ contexts. The mean running time was evaluated
on a simple computer with a quad core processor and $6$ GB of RAM.}
\begin{tabular}{|c|c|c|c|} 
\hline 

Algorithm &  Regret& Classification rate &  Running time\\ 
\hline  \hline
\multicolumn{4}{l}{{\it Forest Cover Type}, action: Cover Type ($7$ types)} \\
\hline
{\sc Banditron} &$1.99$ $10^6$ $\pm 10^5$ &$49.1 \%$ & $10$ min\\
{\sc LinUCB} & $1.23$ $10^6$ $\pm 10^3$ & $60 \%$ & $360$ min\\
{\sc NeuralBandit} & $0.567$ $10^6$ $\pm 2.10^4$ & $68 \%$ & $150$ min\\
\hline
{\sc Bandit Tree} $D=8$ & $0.843$ $10^6$ $\pm 2.10^5$ & $64.2\%$ & $5$ min\\
\hline 
{\sc Bandit Forest} &  & &\\
$D 10-18$ & $0.742$ $10^6$  $\pm 5.10^4$ & $65.8\%$ & $500$ min\\
\hline
\multicolumn{4}{l}{{\it Adult}, action: occupation ($14$ types)} \\
\hline 
{\sc Banditron} &$1.94$ $10^6$ $\pm 3.10^4$ &$21.1 \%$ & $20$ min\\
{\sc LinUCB} & $1.51$ $10^6$ $\pm 4.10^4$ & $25.7 \%$ & $400$ min\\
{\sc NeuralBandit} & $1.2$ $10^6$ $\pm 10^5$ & $29.6 \%$ & $140$ min\\
\hline
{\sc Bandit Tree} $D=8$ & $1.33$ $10^6$ $\pm 10^5$ & $27.9\%$ & $4$ min\\
\hline 
{\sc Bandit Forest} &  & &\\
$D 10-18$ & $1.12$ $10^6$ $\pm 7.10^4$ & $31 \%$ & $400$ min\\
\hline 
\multicolumn{4}{l}{{\it Census1990}, action: Yearsch ($18$ types)} \\
\hline 
{\sc Banditron} &$2.07$ $10^6$ $\pm 2.10^5$ &$27 \%$ & $26$ min\\
{\sc LinUCB} & $0.77$ $10^6$ $\pm 5.10^4$ & $40.3 \%$ & $1080$ min\\
{\sc NeuralBandit} & $0.838$ $10^6$ $\pm 10^5$ & $41.7 \%$ & $300$ min\\
\hline
{\sc Bandit Tree} $D=8$ & $0.78$ $10^6$ $\pm 2.10^5$  & $41\%$ & $10$ min\\
\hline 
{\sc Bandit Forest} &  & &\\
 $D 10-18$ & $0.686$ $10^6$ $\pm 5.10^4$ & $43.2 \%$ & $1000$ min\\
\hline 
\end{tabular} 
\label{tab}
\end {table}
\end {center}

\bibliographystyle{spmpsci}  

\end {document}